\documentclass[a4paper,twoside]{article}

\usepackage{epsfig}
\usepackage{subcaption}
\usepackage{calc}
\usepackage{amssymb}
\usepackage{amstext}
\usepackage{amsmath}
\usepackage{amsthm}
\usepackage{multicol}
\usepackage{pslatex}
\usepackage{apalike}
\usepackage{SCITEPRESS}     
\usepackage{hyperref}
\usepackage{multirow}

\begin{document}

\title{Benchmark for Generic Product Detection: A Low Data Baseline for Dense Object Detection}

\author{\authorname{Srikrishna Varadarajan, Sonaal Kant and Muktabh Mayank Srivastava}
\affiliation{ParallelDots, Inc.}
\email{\{srikrishna,sonaal,muktabh\}@paralleldots.com}
}



\abstract{Object detection in densely packed scenes is a new area where standard object detectors fail to train well \cite{sku110k}. Dense object detectors like RetinaNet \cite{Lin_2017} trained on large and dense datasets show great performance. We train a standard object detector on a small, normally packed dataset with data augmentation techniques. This dataset is 265 times smaller than the standard dataset, in terms of number of annotations. This low data baseline achieves satisfactory results (mAP=0.56) at standard IoU of 0.5. We also create a varied benchmark for generic SKU product detection by providing full annotations for multiple public datasets. It can be accessed at this \href{https://github.com/ParallelDots/generic-sku-detection-benchmark}{URL}. We hope that this benchmark helps in building robust detectors that perform reliably across different settings in the wild.}

\onecolumn \maketitle \normalsize \setcounter{footnote}{0} \vfill

\section{Introduction}

\begin{table*}
\centering
\begin{tabular}{|l|l|l|l|l|l|l|}
\hline
Dataset                 & \#Images & \#Objects & \#Obj/Img & \multicolumn{2}{l|}{Object Size} & Avg Img Size\\ 
                        & & & &   (Mean) & (Std) & \\ \hline
SKU110K-Test            & 2941 & 432,312 & 146 & 0.27\% & 0.21\% & 7.96 \\ \hline
WebMarket               & 3153 & 118,388 & 37 & 1.20\% & 1.09\% & 4.40 \\ \hline
TobaccoShelves          & 354  & 13,184 & 37 & 1.1\% & 0.65\% & 6.08 \\ \hline
Holoselecta             & 295  & 10,036 & 34 & 0.99\% & 0.80\% & 15.62 \\ \hline
GP                      & 680  & 9184 & 13 & 3.66\% & 2.59\% & 7.99 \\ \hline
CAPG-GP                 & 234  & 4756 & 20 & 3.09\% & 3.04\% & 12.19 \\ \hline
\end{tabular}
\caption{Details of the datasets in the benchmark. \# represents the count. Object sizes (Mean and Standard Deviation) are relative to the image size. Average Image size is shown in Megapixels}
\label{table:datasets}
\end{table*}

The real-world applications of computer vision span multiple industries like banking, agriculture, governance, healthcare, automotive, retail, and manufacturing. A few prominent ones include self-driving cars, automated retail stores like Amazon Go, and automated surveillance. The use of object detectors is absolutely a critical part of such real-world products. The area of research in object detectors has been quite vibrant, with a considerable number of datasets spanning various domains. However, the sub-topic of object detection in dense scenes is rarely explored. Dense object detection is quite relevant to multiple applications, for example, in surveillance and retail industries. Some of these applications are crowd counting, monitoring and auditing of retail shelves, insights into brand presence for sales, marketing teams, and so on. 

Exemplar based object detection refers to the detection and classification of objects from scene images with the supervision of an exemplar image of the object. Most object detection datasets are quite large, with enough number of instances of every object category. Most of the object detection methods \cite{fasterrcnn} depend on balanced and large object detection datasets to perform well in every category. These guarantees cannot be made for real-world applications where the object categories vary widely both in variety and in availability. For example, in the retail domain, the gathering of data to train an end to end object detection model is highly time-consuming as well as costly. This is because gathering enough data which covers all the variants of objects and has equal representation of each object is going to be much harder. For example, making sure that our dataset contains a specific rare Mercedes logo design requires us to search across multiple showrooms or market places. We cannot collect a balanced dataset for object detection due to the availability of various logos. A similar case can be made when we need to monitor retail shelves, which has thousands of SKUs having different availability and frequency.

Moreover, in a dynamic world, where new products, new marketing materials, new logos keep getting introduced, the importance of incorporating incremental learning in real-world applications becomes greatly necessary. Unfortunately, the methods of incremental learning for object detection lead to a vast and unacceptable drop in performance \cite{incremental}. A lot of these applications also involve distinguishing between extremely fine-grained classes. E.g., retail shelf monitoring, logo monitoring, face recognition. Building an end-to-end detector that would do both the dense object detection and fine-grained recognition is a very challenging task whose real-world performance is quite bad. Hence, to tackle this problem, we introduce to decouple detection and classification. We propose to use a general object detector that predicts bounding boxes of objects that is of interest. The detected objects can then be classified by a suitable fine-grained classifier. 

This brings us to the current work of generic object detection in densely packed scenes. Previous works \cite{sku110k} have shown that training dense object detectors like RetinaNet \cite{Lin_2017} on large dense object detection datasets works well. In this work, we explore the effectiveness of standard object detectors, trained on very low data. Our training data is 265 times smaller than the other methods, in terms of number of annotations.  We provide a low data baseline for dense object detection task. We train a standard detector, namely Faster-RCNN \cite{fasterrcnn}, on a small dataset of normal scenes. This achieves a satisfactory performance (mAP) at the standard IoU of 0.5.

We also create a varied benchmark for generic SKU product detection by annotating every SKU in multiple datasets. The motivation behind this is to create detectors that are robust across different settings. It is quite common for deep learning based detectors to transfer poorly to other datasets. In the industry, there is a high need for robust detectors that perform reliably in the wild. This benchmark consists of 6 datasets (Table \ref{table:datasets}) used solely as test sets. Models trained on any dataset can be tested on this benchmark to measure the progress of robust generic SKU detection. The benchmark datasets, evaluation code, and the leaderboard are available at this URL\footnote{https://github.com/ParallelDots/generic-sku-detection-benchmark}

\section{Related Work}

Focal loss for dense object detection was introduced by \cite{Lin_2017}. A method to localize objects more precisely in dense scenes was proposed by \cite{sku110k}. This achieves significant increase in performance (mAP) at higher IoUs. Identifying real-world products (in situ) by training on exemplar template product images (in vitro) was initially proposed by \cite{grozi}. They released a database of 120 SKUs for product classification. Six in vitro images were collected from the web for each product and used for training.  The in situ images were provided from frames of videos captured in a grocery store. There have been a few retail product checkout datasets by \cite{RPC} and \cite{MVTecD2S}. Both of them are densely packed product datasets arranged in the fashion of a checkout counter, with many overlapping regions between objects as well. 


\begin{table*}
\centering
\begin{tabular}{|l|l|l|l|l|l|l|l|l|}
\hline
\multirow{2}{*}{Dataset} & \multirow{2}{*}{\#Images} & \multicolumn{7}{l|}{\#Images with SKU count greater than} \\ \cline{3-9} 
                         &                           & 5    & 10   & 20       & 40      & 60    & 80   & 100      \\ \hline
SKU110K-Test             & 2941                      & 2941 & 2941 & 2939     & 2934    & 2926  & 2902 & 2822     \\ \hline
WebMarket                & 3153                      & 3139 & 3048 & 2526     & 1119    & 349   & 121  & 57     \\ \hline
TobaccoShelves           & 354                       & 354  & 349  & 308      & 130     & 18    & 3    & 2     \\ \hline
Holoselecta              & 295                       & 293  & 293  & 249      & 113     & 0     & 0    & 0    \\ \hline
GP                       & 680                       & 604  & 407  & 101      & 9       & 1     & 0    & 0     \\ \hline
CAPG-GP                  & 234                       & 209  & 160  & 82       & 33      & 4     & 0    & 0    \\ \hline    
\end{tabular}
\caption{Analysis on the denseness of the generic product datasets. \# represents the count.}
\label{table:denseness}
\end{table*}

\section{Benchmark Datasets}

\cite{sku110k} recently released a huge benchmark dataset for product detection in densely packed scenes. To increase diversity to the task of generic product detection, we release a benchmark of datasets. Details of the datasets are shown in Table \ref{table:datasets}.  Please note that all of these datasets are used as a test set on which we benchmark our models. We welcome the community to participate in this benchmark by submitting their results. 

\subsection{WebMarket}
\cite{WebMarket} released a database of 3153 supermarket images. They also provided information regarding what product is present in each image. We annotate every object in the entire dataset to provide ground truth for the evaluation of general object detection.  The average number of objects per image in this dataset is 37, while the average object area is roughly 0.052 megapixels.

\subsection{Grocery products (GP)}
A multi-label classification approach was proposed by \cite{GP} accompanied by 680 annotated shop images from their GP dataset. The annotation provided by them covered a group of same products in a bounding box rather than bounding boxes for individual boxes. A subset of 70 images was chosen by \cite{GP_some_annotations} and annotated with the desired object-level annotations. We provide individual bounding box annotation for every product for all 680 images in this dataset. The average number of objects per image in this dataset is 13, while the average object size is roughly 0.293 megapixels.

\subsection{CAPG-GP}
A fine-grained grocery product dataset was released by \cite{FGC-OSL}. It consists of 234 test images taken from 2 stores. The authors annotated only the products belonging to certain categories. To create ground truth for generic object detection, we decided to annotate every product in the entire dataset. The average number of objects per image in this dataset is 20, while the average object area is roughly 0.377 megapixels.

\begin{table}
\centering
\begin{tabular}{|l|l|l|l|l|}
\hline
Type of Images          & Avg Img Size & \#Images           \\ \hline
Type 1 (HoloLens)               & 1.11  & 30 \\ \hline
Type 2                          & 8.13  & 8 \\ \hline
Type 3 (OnePlus-6T)             & 16.03 & 208 \\ \hline
Type 4                          & 24    & 49 \\ \hline
Total                           & 15.62 & 295 \\ \hline
\end{tabular}
\caption{Details of Holoselecta dataset. \# represents the count. Average size of the entire image is shown in megapixels}
\label{table:holoselecta}
\end{table}

\subsection{Existing General Product Datasets}

\subsubsection{Holoselecta}
Most recently, \cite{holoselecta} released a dataset of 300 real-world images of Selecta vending machines containing 10,000 objects belonging to 90 categories of packaged products. The images in this dataset were quite varied in their sizes, as shown in Table \ref{table:holoselecta}.

\subsubsection{TobaccoShelves}
A retail product dataset was released by \cite{TurkeyCigarette} containing 345 images of tobacco shelves collected 40 stores with four cameras. The annotations of every product were also released by the authors. The average number of objects per image in this dataset is 37, while the average object area is roughly 1.1\% of the entire image. The images in this dataset were also quite varied in size, ranging from 1.4 megapixels to 10.5 megapixels.

\subsubsection{SKU110K}
Recently, \cite{sku110k} released a huge dataset for precise object detection in densely packed scenes. This dataset contains 11,762 images that were split into train, validation, and test sets. The test set consists of 2941 images with the average number of objects per image being 146. The average object area is roughly 0.27\%, making it the lowest among all the datasets in this benchmark.

\subsection{Denseness of the Datasets}
The denseness of the datasets depends on two factors. The average number of objects and the relative sizes of the objects. SKU110K \cite{sku110k} is by far one of the most dense datasets for object detection. An analysis of the denseness of other datasets in the current benchmark is shown in Table \ref{table:denseness}.

\begin{table}
\centering
\begin{tabular}{|l|l|l|l|l|}
\hline
Dataset       & \#Images & \#Obj/Img & \#Anns \\ \hline
Our trainset  & 312      & 14.6      & 4556           \\ \hline
SKU110K-train & 8233     & 147.4     & 1,210,431        \\ \hline
\end{tabular}
\caption{Statistics of trainset. \# represents the count. Number of Annotations is denoted by \textit{\#Anns}}
\label{trainingset}
\end{table}

\begin{table*}
\centering
\begin{tabular}{|l|l|l|l|l|l|l|}
\hline
Dataset                & Method     & AP    & AP$^{.50}$ & AP$^{.75}$ & AR$_{300}$ & AR$_{300}^{.50}$  \\ \hline
SKU110K-Test           & RetinaNet \cite{sku110k} & 0.455 & - & 0.389 & 0.530 & - \\ 
                       & Full Approach \cite{sku110k} & 0.492 & - & 0.556 & 0.554 & - \\ 
                       & Full Approach* & 0.514 & 0.853 & 0.569 & 0.571 & 0.872 \\ 
                       & Faster-RCNN \cite{sku110k} & 0.045 & - & 0.010 & 0.066 & - \\ 
                       & \emph{LDB300}  & 0.186 & \textbf{0.560} & 0.052 & 0.264 & 0.647 \\ \hline
WebMarket              & Full Approach* & 0.383 & 0.773 & 0.332 & 0.491 & 0.855  \\ 
                       & \emph{LDB300}  & \textbf{0.322} & 0.621 & 0.248 & 0.455 & 0.684  \\ \hline
TobaccoShelves         & Full Approach* & 0.534 & 0.948 & 0.560 & 0.615 & 0.970  \\ 
                       & \emph{LDB300}  & 0.108 & 0.442 & 0.009 & 0.159 & 0.491  \\ \hline
Holoselecta            & Full Approach* & 0.454 & 0.835 & 0.447 & 0.581 & 0.955  \\ 
                       & \emph{LDB300}  & 0.239 & 0.707 & 0.072 & 0.347 & 0.816  \\ \hline
GP                     & Full Approach* & 0.259 & 0.520 & 0.241 & 0.403 & 0.716  \\ 
                       & \emph{LDB300}  & \textbf{0.234} & \textbf{0.596} & 0.125 & 0.334 & 0.713  \\ \hline
CAPG-GP                & Full Approach* & 0.431 & 0.684 & 0.519 & 0.481 & 0.721  \\
                       & \emph{LDB300}  & 0.312 & \textbf{0.745} & 0.169 & 0.434 & 0.895  \\ \hline
\end{tabular}
\caption{Performance of $our$ Faster-RCNN across different general product datasets. * denotes results obtained using the trained model given at \href{https://github.com/eg4000/SKU110K_CVPR19/issues/9}{URL} as is.}
\label{table:results}
\end{table*}

\section{Low Data Baseline Approach}
We collected close to 300 images encompassing various shapes in which retail products occur by querying images from the public domain (e.g., GoogleImages, OpenImages). The total number of annotations in our dataset is 4556. This is in contrast with SKU110K-Train, the data on which \cite{sku110k} was trained, where the total number of annotations is 1.2 million, as shown in Table \ref{trainingset}. This makes our training dataset 265 times smaller than the SKU110K-Train. 

We apply standard object detection augmentations from \cite{albumentations}. We train a standard object detector, Faster-RCNN \cite{fasterrcnn}, on our training set described above as a low data baseline for this benchmark. We call this model \emph{LDB300} (Low Data Baseline). We measure the performance across different datasets, to test the robustness of the model in the wild and its effectiveness in generic product detection.

\section{Implementation \& Results}
We use standard post-processing steps like Non-Max Suppression after our detections. We use multi-scale testing (2 scales) since lot of the datasets have high variance in object sizes. For the SKU110K dataset, only one scale is used. The inference settings of the compared \emph{Full Approach} can be obtained from the code base released by the authors at \href{https://github.com/eg4000/SKU110K_CVPR19}{URL}. Multi-scale testing can be employed on the Full Approach as well, but this might be highly time consuming. For example, it is 10 times slower (FPS) than Faster-RCNN, as shown in \cite{sku110k}.

We use the same evaluation metric as the recent work \cite{sku110k}. This is the standard evaluation metric used by COCO \cite{COCO}. Average precision (AP) and  Average Recall (AR$_{300}$) are reported at IoU=.50:.05:.95 (averaged by varying IoU between 0.5 and 0.95 with 0.05 intervals). 300 here represents the maximum number of objects in an image. AP and AR at specific IoUs (0.50 and 0.75) are also reported.

Our baseline results on the SKU110K dataset (Table \ref{table:results}) shows satisfactory performance of mAP=0.560 at 0.5 IoU, even with very low data (265 times in terms of number of annotations). For some of the datasets like GP, CAPG-GP, our baseline results are quite close to the state-of-the-art. This could be because of the huge variation in object sizes as well the non-dense nature of the scenes in these datasets, which can be seen from Table \ref{table:datasets} and Table \ref{table:denseness}. This method serves as a simple baseline while methods exploiting the shape of the objects and structure of densely packed scenes look promising.

\section{Discussion}
The varying results from Table \ref{table:results} shows that evaluating on a varied benchmark is necessary to have a detector perform reliably in the wild. For example, the Full Approach \cite{sku110k} trained on the huge SKU110K dataset does not perform well on the GP \cite{GP} dataset as is.

\begin{figure*}
\includegraphics[width=0.5\textwidth,height=0.4\textheight]{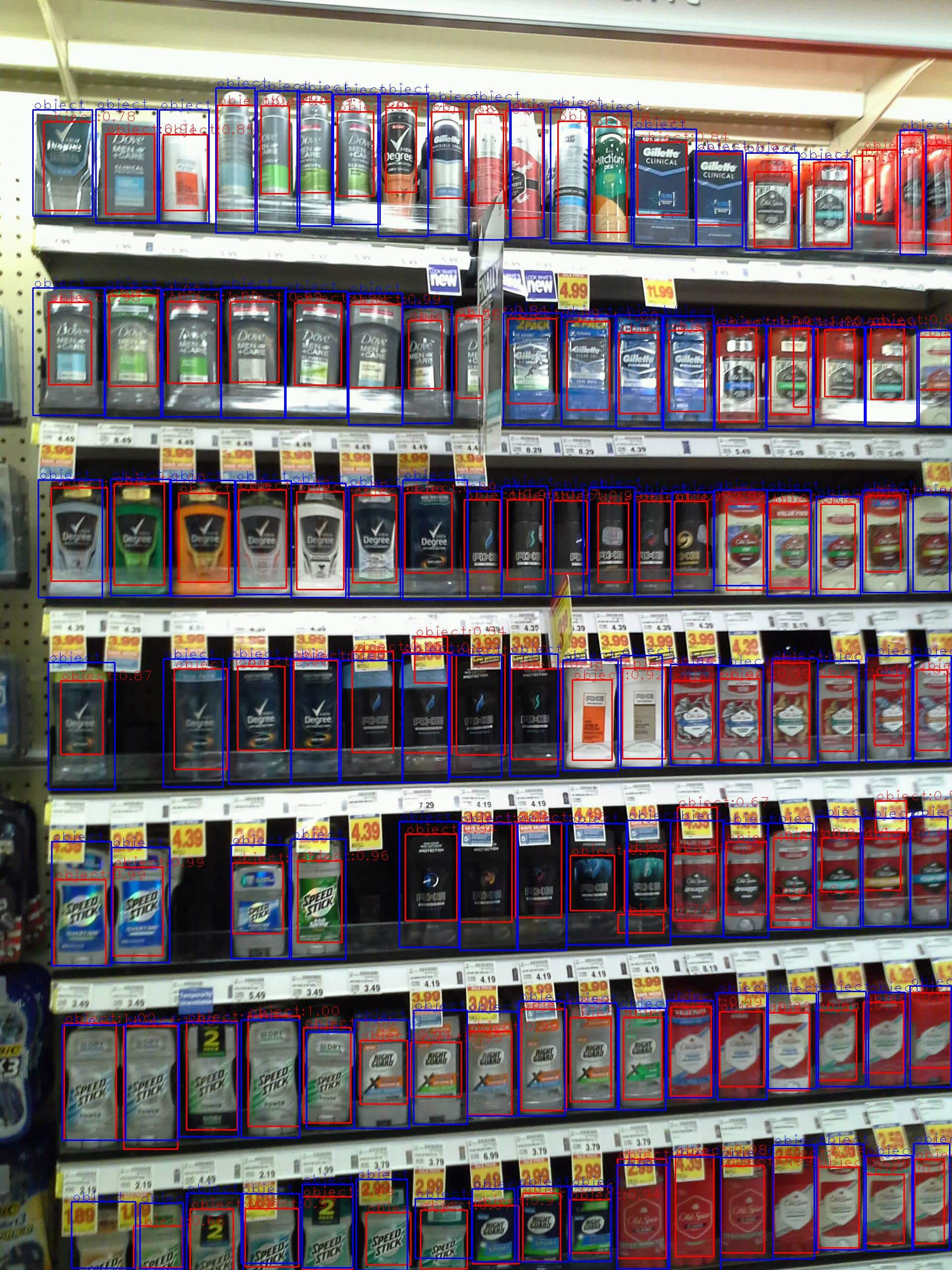}
\includegraphics[width=0.5\textwidth,height=0.4\textheight]{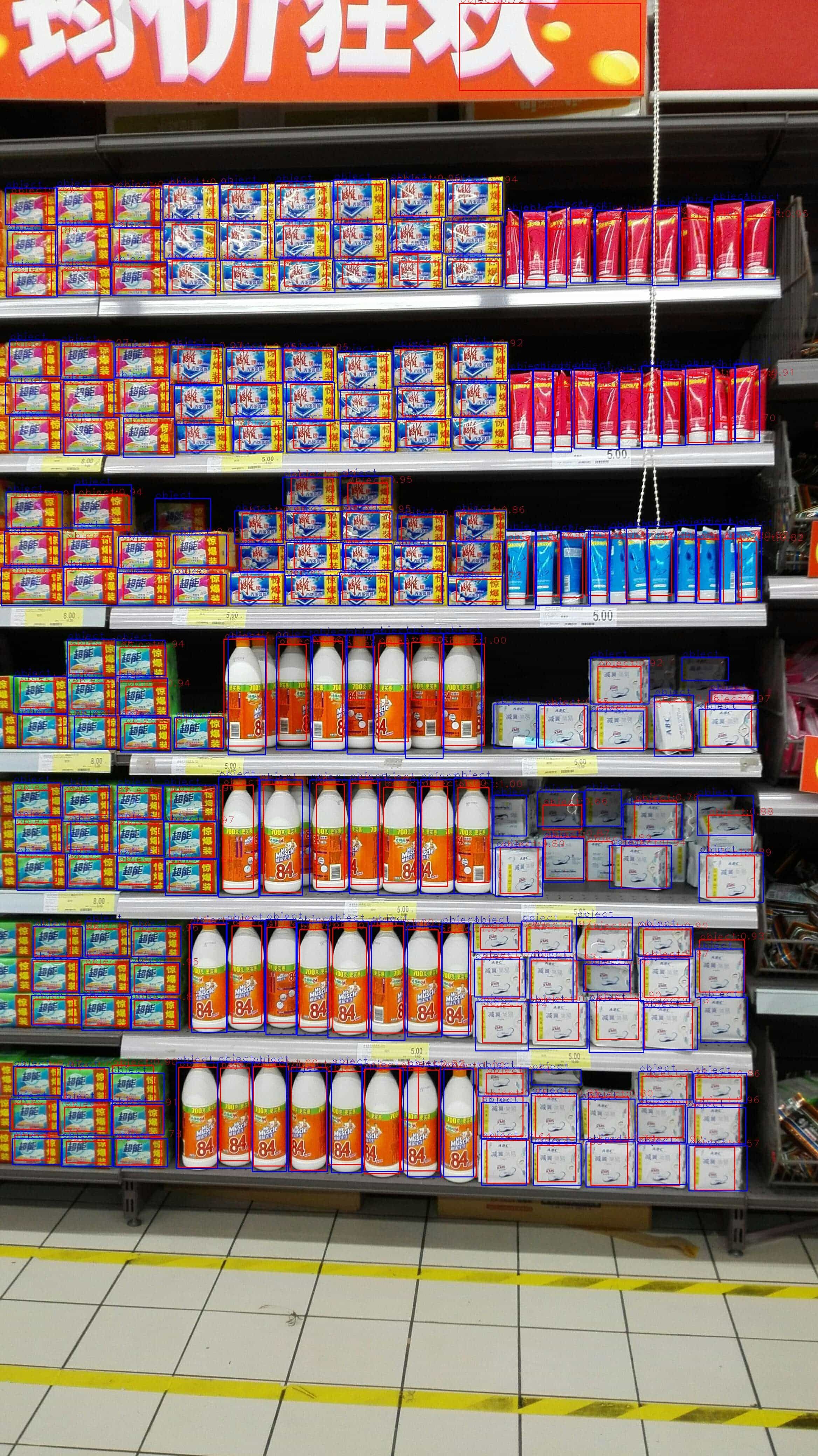}
\includegraphics[width=0.5\textwidth,height=0.4\textheight]{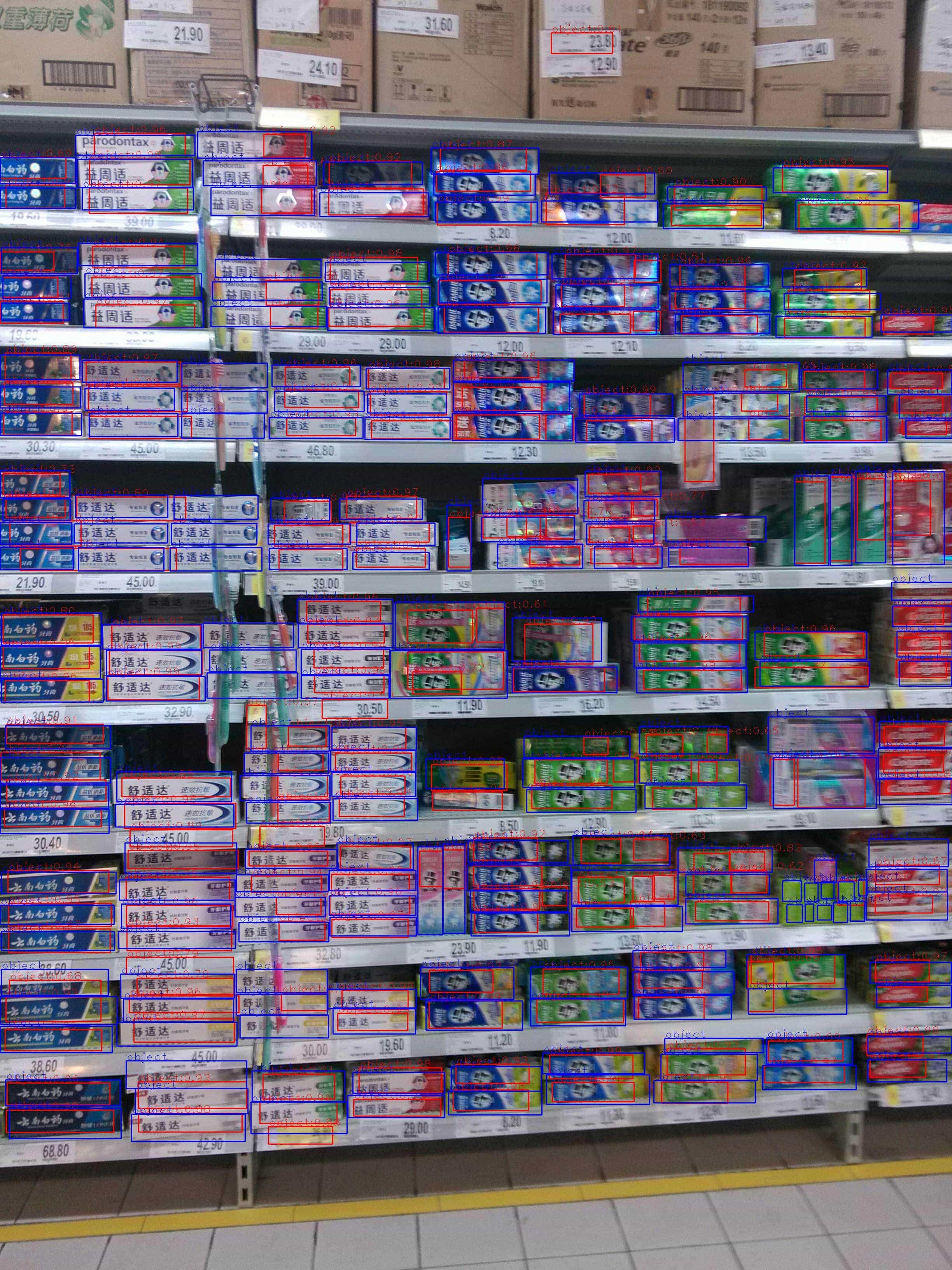}
\includegraphics[width=0.5\textwidth,height=0.4\textheight]{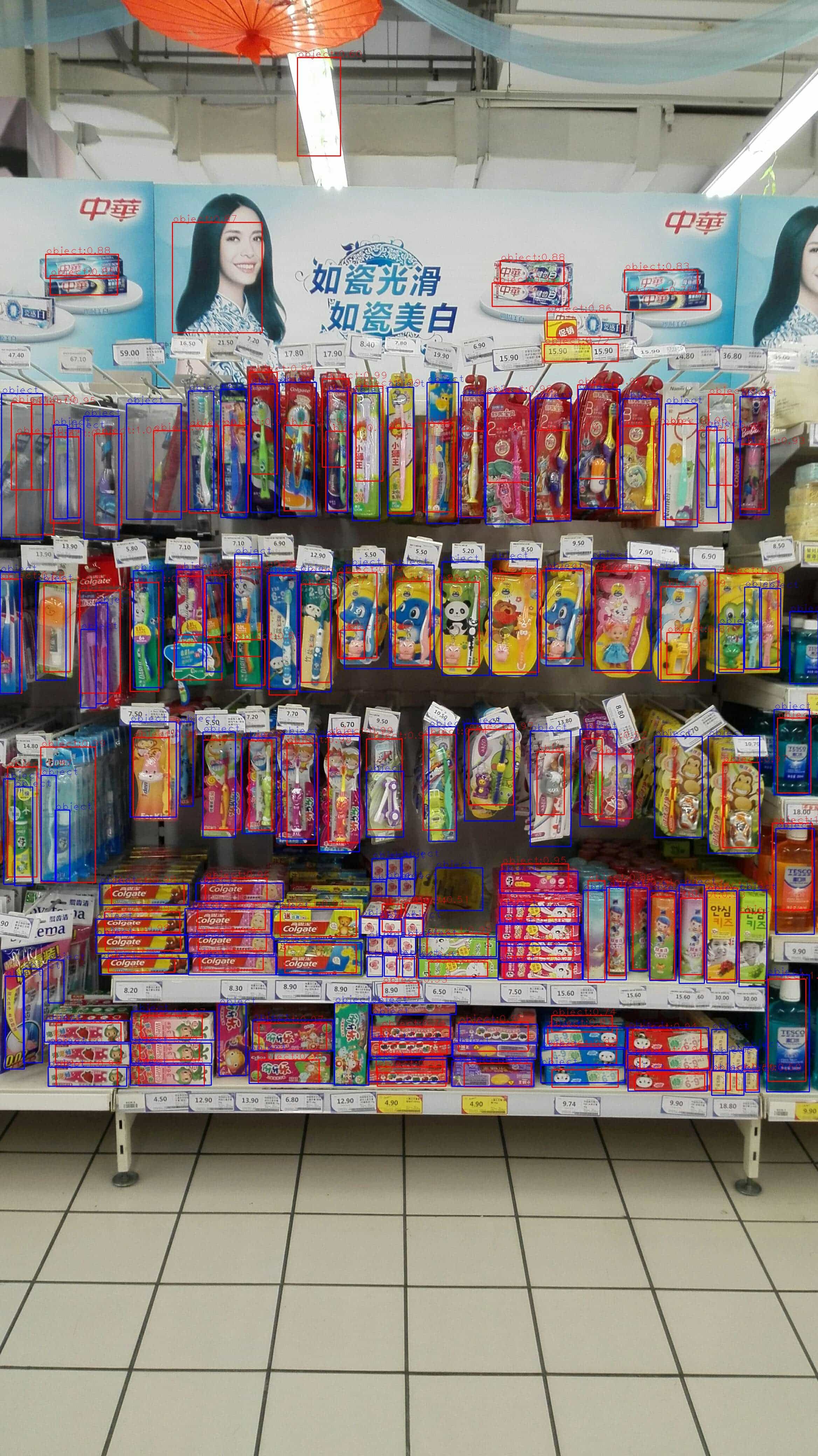}
  \caption{Ouputs of our method spanning different type of objects on SKU110K \cite{sku110k} dataset.}
\label{fig:outputs-sku}
\end{figure*}

\begin{figure*}
\includegraphics[width=0.5\textwidth,height=0.3\textheight]{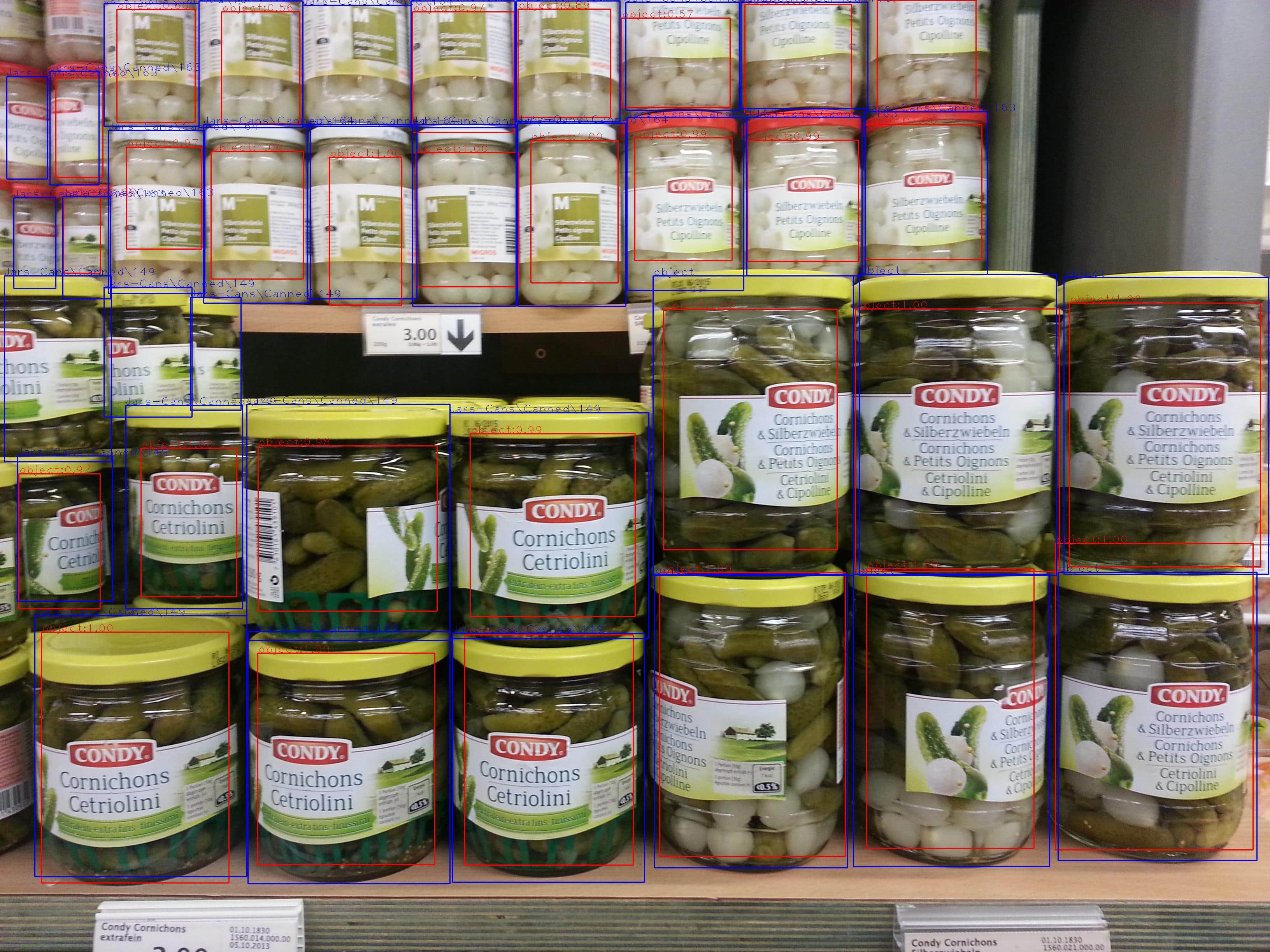}
\includegraphics[width=0.5\textwidth,height=0.3\textheight]{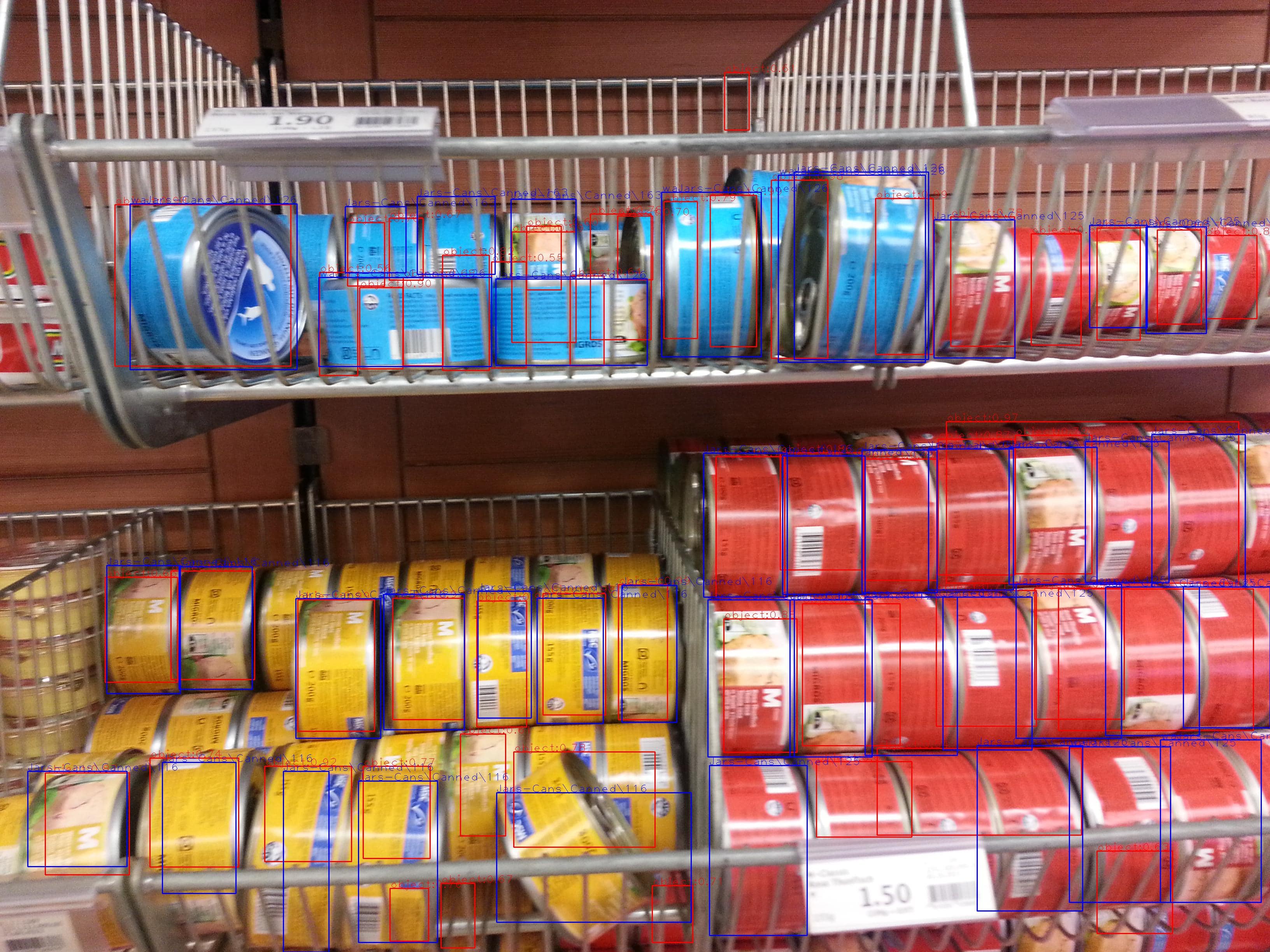}
  \caption{Sample predictions of our method on GP \cite{GP} dataset. Blue boxes denote groundtruth objects, while Red denotes predictions.}
\label{fig:outputs-GP}
\end{figure*}

\begin{figure*}
\includegraphics[width=0.5\textwidth,height=0.3\textheight]{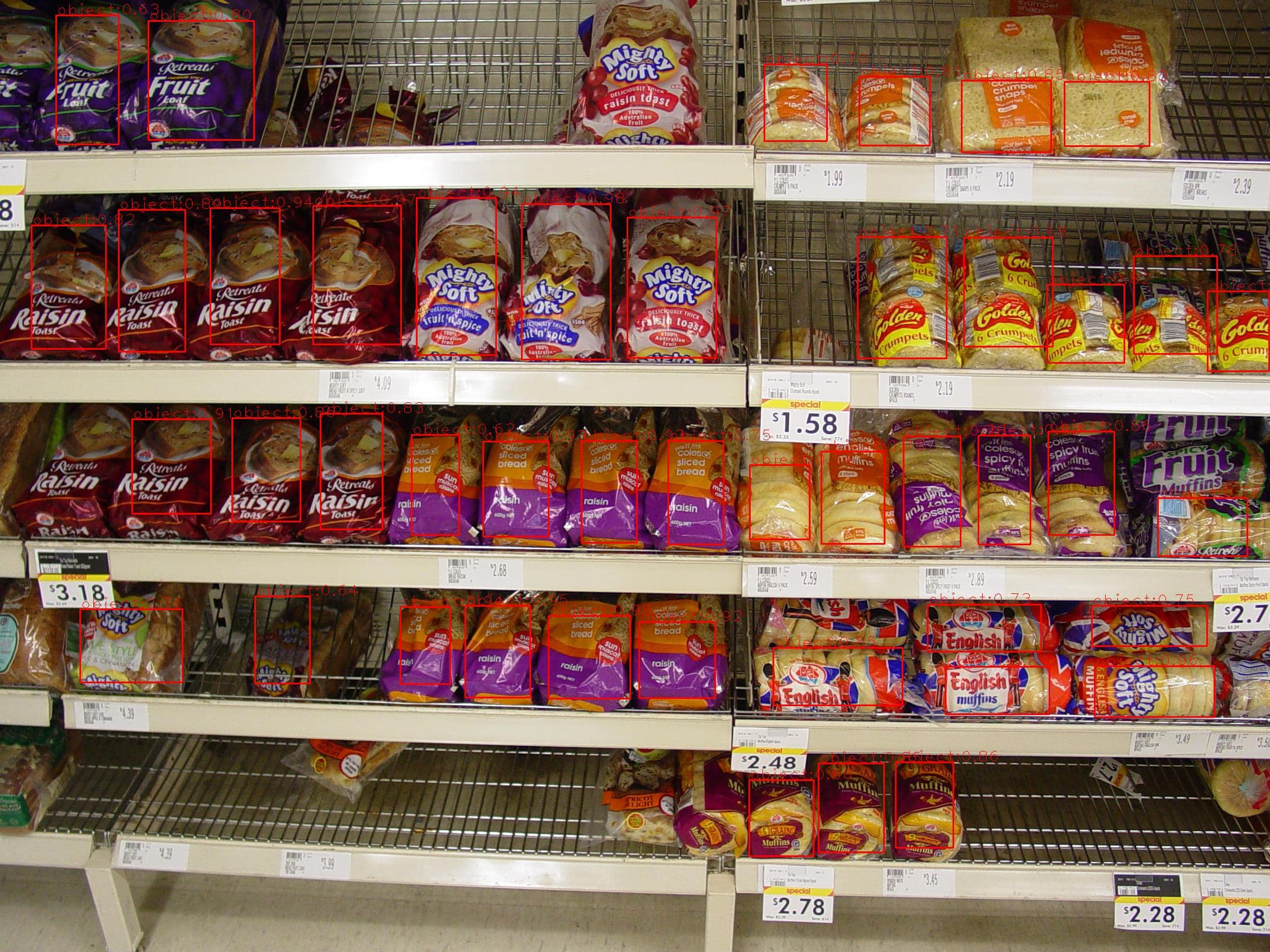}
\includegraphics[width=0.5\textwidth,height=0.3\textheight]{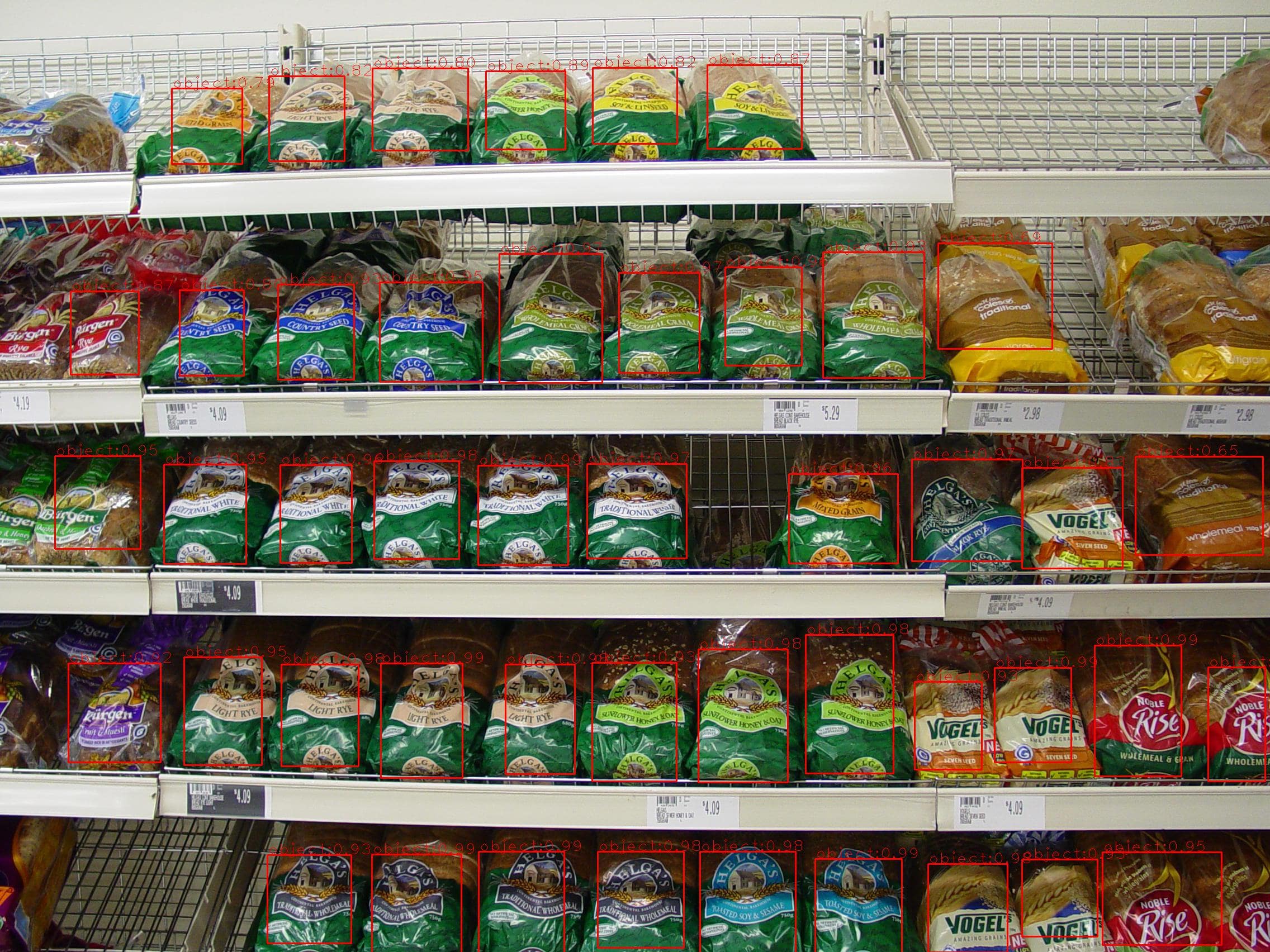}
  \caption{Ouputs of our method on Holoselecta \cite{WebMarket} dataset}
\label{fig:outputs-WebMarket}
\end{figure*}

\begin{figure*}
\includegraphics[width=0.5\textwidth,height=0.3\textheight]{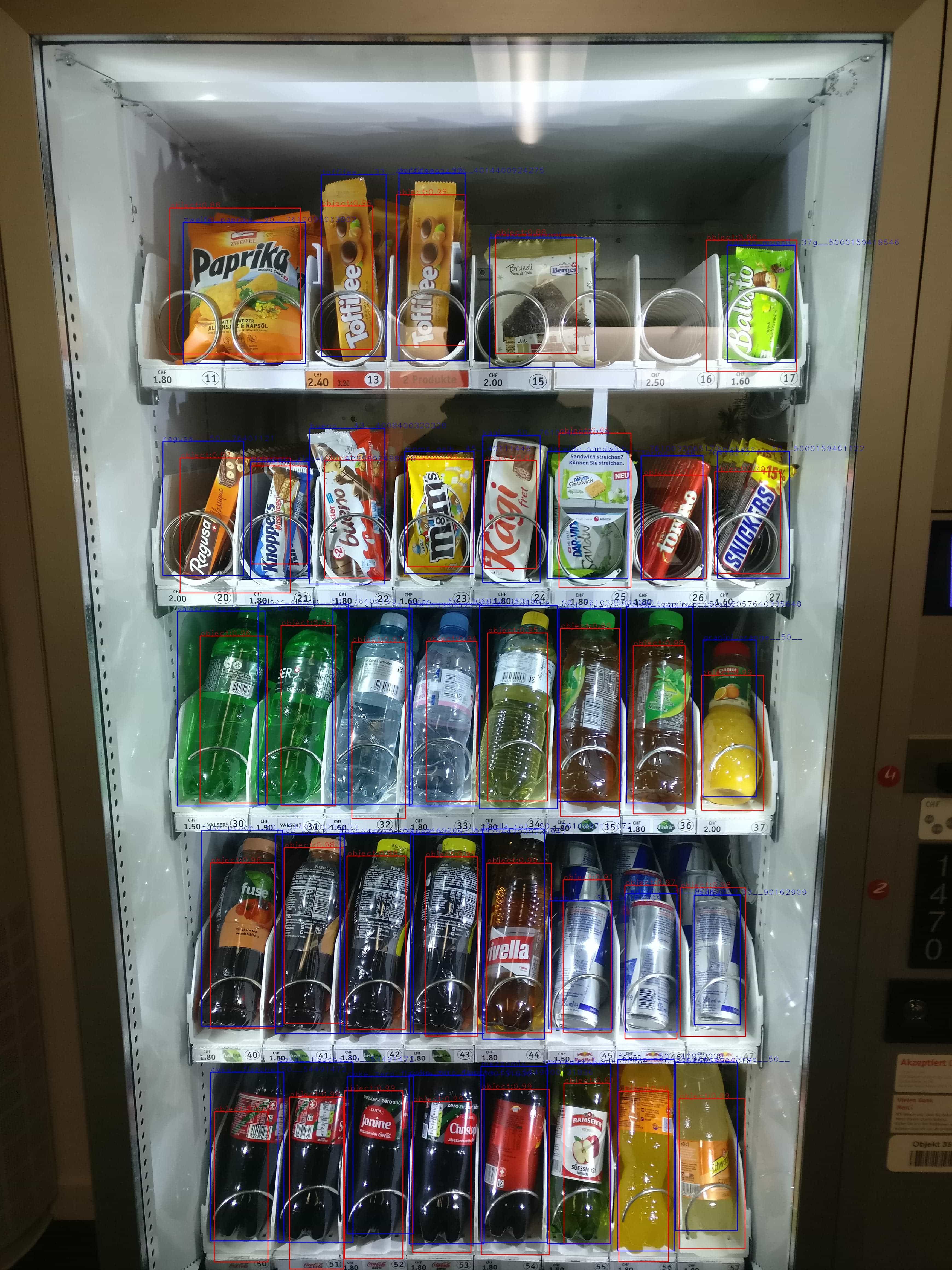}
\includegraphics[width=0.5\textwidth,height=0.3\textheight]{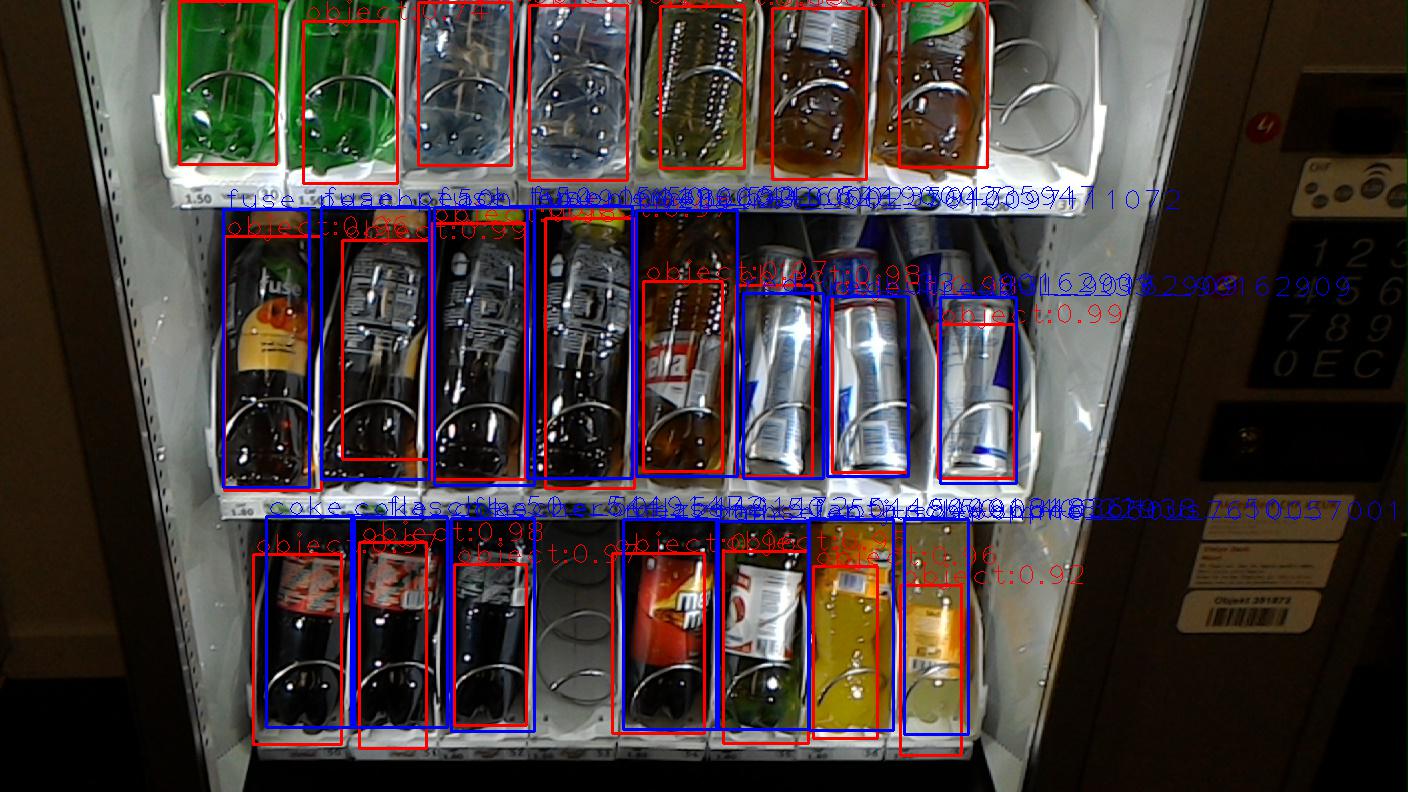}
  \caption{Ouputs of our method on Holoselecta \cite{holoselecta} dataset}
\label{fig:outputs-Holoselecta}
\end{figure*}

\begin{figure*}

\includegraphics[width=0.5\textwidth,height=0.4\textheight]{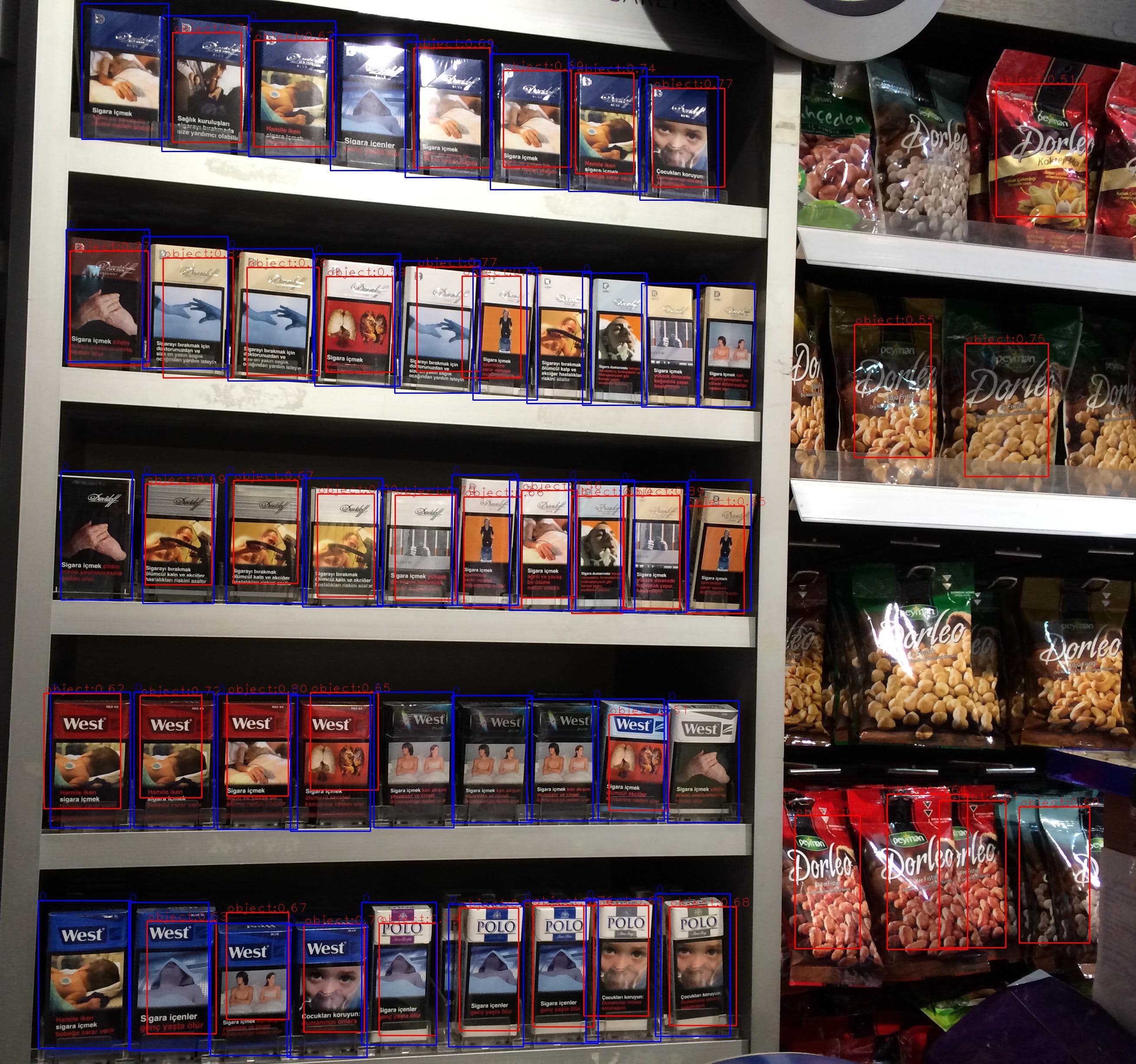}
\includegraphics[width=0.5\textwidth,height=0.2\textheight]{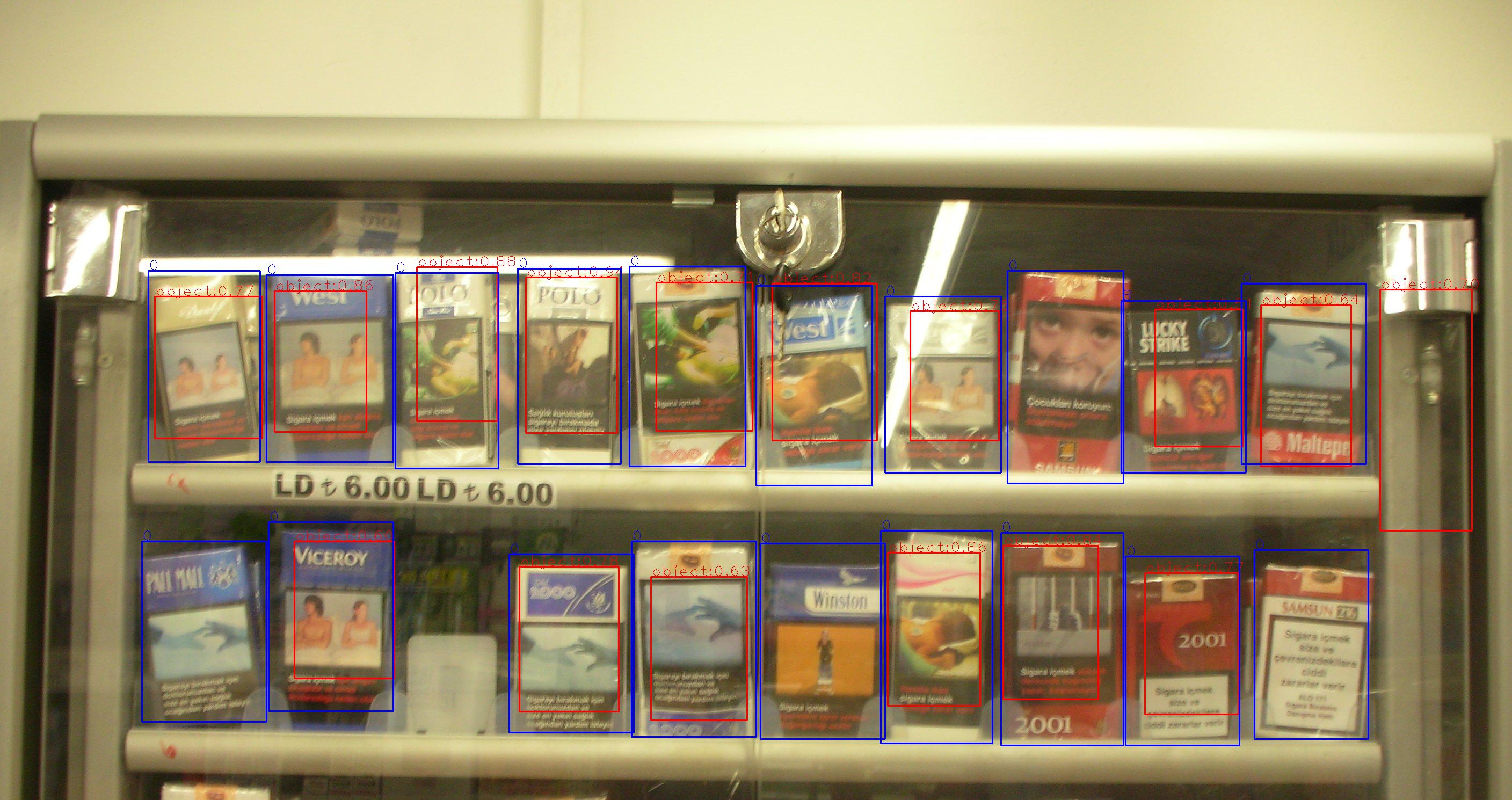}
\begin{center}
\includegraphics[width=0.7\textwidth,height=0.4\textheight]{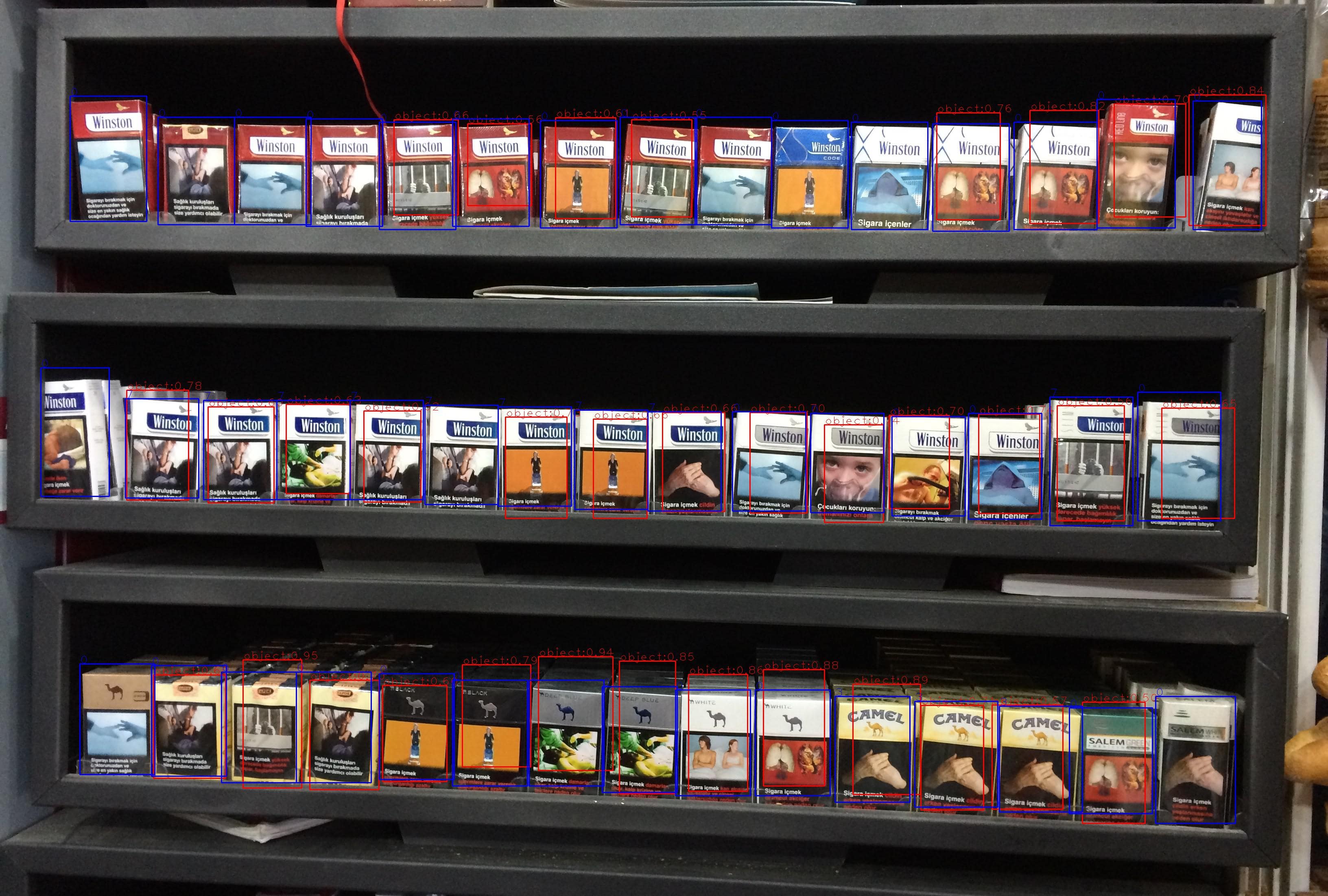}
\end{center}
  \caption{Ouputs of our method on TobaccoShelves \cite{TurkeyCigarette} dataset}
\label{fig:outputs-tc}
\end{figure*}

A qualitative output of our method on different datasets are shown in Figure \ref{fig:outputs-sku}, \ref{fig:outputs-GP}, \ref{fig:outputs-WebMarket}, \ref{fig:outputs-Holoselecta}, \ref{fig:outputs-tc}. The performance on TobaccoShelves was a bit low (Table \ref{table:results}), which is seen qualitatively in Figure \ref{fig:outputs-tc}. One can see that, our method performs well when homogeneous objects are present throughout the image. It also detects objects of different aspect ratios comfortably. Current limitations of this baseline model include precise detection of the objects, detecting objects with occlusion from shelves, as well as handling multi-scale objects ranging from 0.1\% to 20\% of the scene. Precise detection can be helped by EM-Merger module from \cite{sku110k}. Better scale-invariant object detectors \cite{snip} can be tried as future work.

\bibliographystyle{apalike}
{\small
\bibliography{example}}

\end{document}